%% file: main.tex
\documentclass[10pt,twocolumn,letterpaper]{article}

\usepackage{cvpr}              % To produce the CAMERA-READY version
	
\usepackage{color, colortbl}
\definecolor{lightgreen}{rgb}{0.8,1.0,0.9}
\definecolor{lightblue}{rgb}{0.8,0.9,1.0}
\definecolor{lightgray}{rgb}{0.9,0.9,0.9}
\definecolor{lightviolet}{rgb}{0.95,0.9,1.0}
\definecolor{lightyellow}{rgb}{1.0,1.0,0.8}

\usepackage[normalem]{ulem}
\usepackage{pifont}

\usepackage[numbers]{natbib}

\definecolor{cvprblue}{rgb}{0.21,0.49,0.74}
\usepackage[pagebackref,breaklinks,colorlinks,allcolors=cvprblue]{hyperref}

\def\paperID{*****} % *** Enter the Paper ID here
\def\confName{CVPR}
\def\confYear{2026}

\title{Training-Free Long-Term Multi-Object Tracking for Sports Video Analytics}

\author{
Tomasz Stanczyk$^{1,2}$ \qquad Seongro Yoon$^{1,2}$ \qquad Francois Bremond$^{1,2}$\\
$^1$ Inria, France  \hspace{0.5cm} $^2$ Université Côte d’Azur, France\\
{\tt\small Corresponding author: Tomasz Stanczyk, tomasz.stanczyk@inria.fr}
}

\begin{document}
%TC:ignore
\maketitle
\begin{abstract}
Long-term multi-object tracking in sports remains challenging due to frequent occlusions, rapid camera motion, and repeated player reappearances. We introduce McByte++, a training-free tracking-by-detection framework that integrates lightweight mask propagation, conditional camera motion compensation, and online re-identification within a unified pipeline. Compared to its predecessor, McByte++ substantially improves runtime efficiency while enhancing identity preservation. On SoccerNet-tracking and SportsMOT benchmarks, McByte++ achieves up to +3.0 HOTA and +6.1 IDF1 improvements over the original McByte in the online setting, with further gains when combined with offline global association. Replacing heavy segmentation components and optimizing motion modeling yields up to an order-of-magnitude speed increase. All results are obtained without detector retraining or dataset-specific tuning. Code will be made available at \url{https://github.com/tstanczyk95/McBytePlusPlus}.
\end{abstract}
%TC:endignore

%%%%%%%%% BODY TEXT
\section{Introduction}
\label{sec:intro}

Multi-object tracking (MOT) is a core component of modern sports video analytics, enabling the extraction of player trajectories, tactical patterns, and performance statistics across a wide range of sports. Unlike generic pedestrian tracking, sports MOT operates under particularly challenging conditions: fast and abrupt motion, frequent occlusions, rapid camera movements, motion blur, and highly dynamic scene layouts~\cite{sportsmot_ref,soccernet-tracking2022_ref,soccernet-tracking2023_ref}. These challenges are further amplified by broadcast-style footage, where players regularly leave and re-enter the field of view due to camera framing, substitutions, or changes in play. As a result, sports tracking systems must be not only accurate, but also robust, efficient, and capable of maintaining identity consistency over extended temporal gaps.

Most contemporary MOT systems rely on training-heavy paradigms, including end-to-end transformer-based trackers or deeply learned association models~\cite{motr_ref,memotr_ref,motrv2_ref,motip_ref,trackformer_ref}. While such approaches have demonstrated strong performance on curated benchmarks, they introduce several limitations in sports settings. First, they require large amounts of annotated training data, which is costly and often unavailable across different sports and leagues. Second, their performance is typically sensitive to dataset bias and domain shift, limiting their generalization across sports, camera setups, and broadcast styles. Finally, their computational complexity often limits their practical deployment for live or large-scale sports analytics, where computational efficiency is an important requirement.

An alternative and widely adopted paradigm is tracking-by-detection, where objects are detected independently in each frame and associated over time using motion and geometric cues~\cite{sort_ref,deepsort_ref,bt_ref,ocsort_ref,strongsort_ref,hybridsort_ref}. Methods in this category are attractive for sports analytics due to their modularity, interpretability, and reliance on off-the-shelf detectors. However, classical tracking-by-detection methods primarily focus on short-term association, often terminating identities once an object disappears for a limited number of frames. In sports scenarios, where players may be occluded for extended periods or temporarily exit the camera view, this behavior leads to fragmented trajectories and a loss of long-term identity consistency.

Long-term tracking, defined here as the ability to reconnect tracked entities after extended absences, remains underexplored in sports MOT~\cite{tracktor_ref, quovadis_ref}. Existing solutions typically address this problem via offline~\cite{sushi_ref} and post-processing~\cite{gta_link_ref}, rather than through online, integrated mechanisms. Moreover, long-term identity association is often introduced at the cost of increased computational overhead or additional training requirements~\cite{corr_learning_ref, strongsort_ref}, limiting its practicality in real-world deployments.

In prior work, we introduced McByte~\cite{MCBYTE_REF}, a training-free tracking-by-detection framework that integrates temporally propagated segmentation masks~\cite{xmem_ref,cutie_ref} as an association cue. By leveraging mask propagation models pretrained on large-scale data, McByte improves robustness to occlusions and motion blur without requiring any task-specific training or per-sequence parameter tuning. While effective in short-term tracking, the original McByte formulation did not explicitly address long-term identity reconnection when objects leave and re-enter the scene.

In this work, we present McByte++, a substantial extension of McByte that advances training-free sports MOT toward long-term identity tracking, while simultaneously improving efficiency and runtime performance. McByte++ introduces an optional, lightweight re-identification (re-ID)~\cite{osnet_ref,market1501_ref} mechanism that operates online, allowing newly created tracklets to be connected to previously observed entities when they reappear after extended temporal gaps. Crucially, this long-term association is performed without additional training and can be enabled either during runtime or as a post-processing refinement step. This design enables McByte++ to handle realistic sports scenarios involving substitutions, camera cuts, and prolonged occlusions - cases that are central to sports analytics but poorly supported by most existing trackers.

Beyond long-term tracking, McByte++ introduces several system-level improvements that make it suitable for practical deployment. We improve camera motion compensation to better handle broadcast footage with abrupt viewpoint changes, and we replace heavy segmentation components with a lightweight mask propagation module, significantly reducing computational cost. As a result, McByte++ achieves substantial gains in both accuracy and computational efficiency, significantly increasing processing speed while consistently improving identity preservation metrics such as IDF1 and HOTA across multiple sports datasets.

The resulting system retains the core philosophy of McByte: no training, no tuning, and strong generalization across sports. McByte++ relies exclusively on existing pretrained detectors, mask propagation models, and re-ID feature extractors, and uses the same detections as competing tracking-by-detection methods to ensure fair comparison. By design, it avoids dataset-specific optimization, making it applicable across different sports, camera setups, and levels of play.

In summary, this paper makes the following contributions:

\begin{itemize}
    \item We extend the original McByte framework from short-term mask-guided association toward efficient long-term multi-object tracking by introducing a unified training-free framework that selectively combines complementary tracking cues.
    
    \item We propose an online long-term identity association mechanism that invokes re-identification only when new tracklets are initialized, enabling reconnection with previously terminated identities while avoiding continuous appearance matching during short-term association.
    
    \item We redesign the tracking pipeline through lightweight mask propagation and conditional camera motion compensation, substantially improving computational efficiency while preserving robust association performance.
    
    \item We demonstrate through extensive experiments on SportsMOT and SoccerNet (2022 and 2023) benchmarks that McByte++ consistently improves identity preservation and long-term tracking performance without detector retraining, dataset-specific tuning, or additional supervision.
\end{itemize}

\begin{figure*}
\centering
\includegraphics
[width=15cm]
{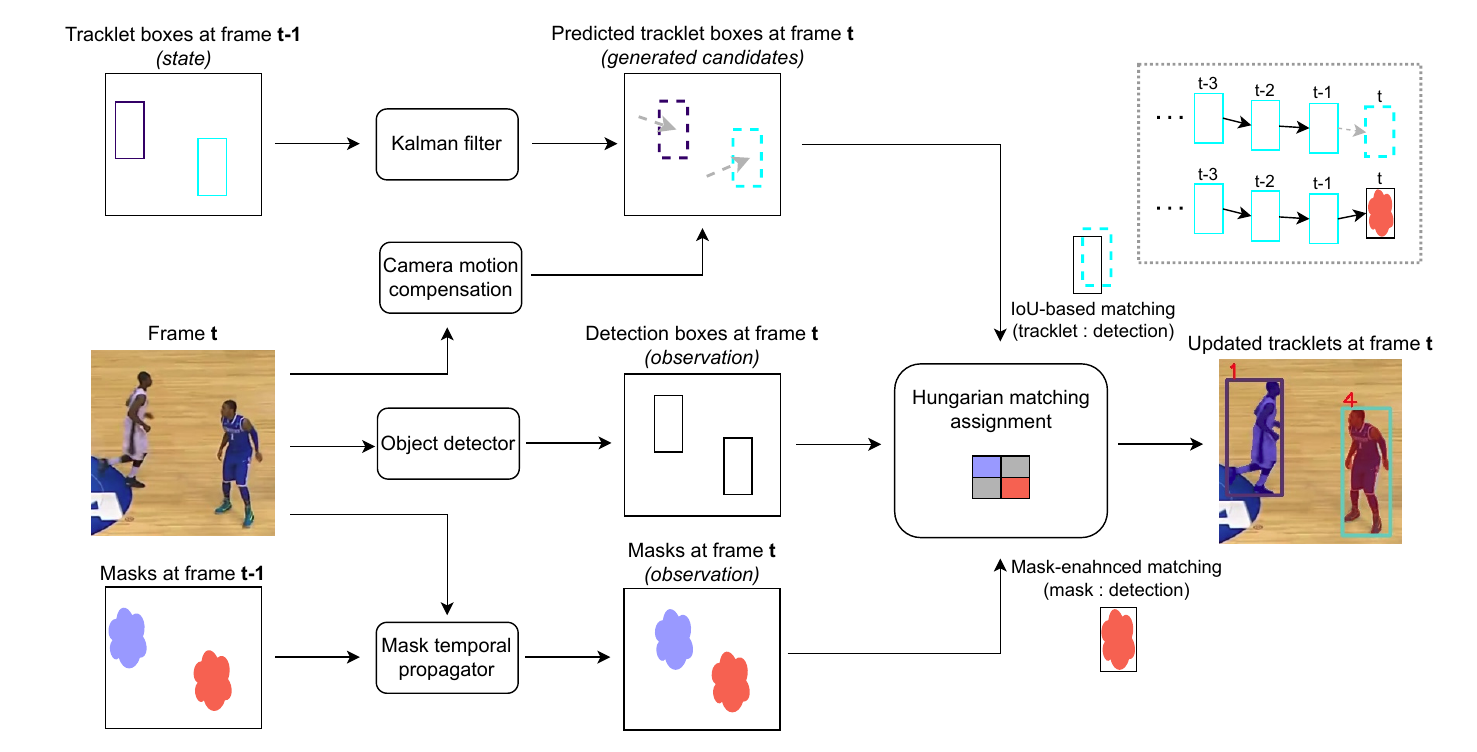}\\
(a) \\
\includegraphics[width=15cm]{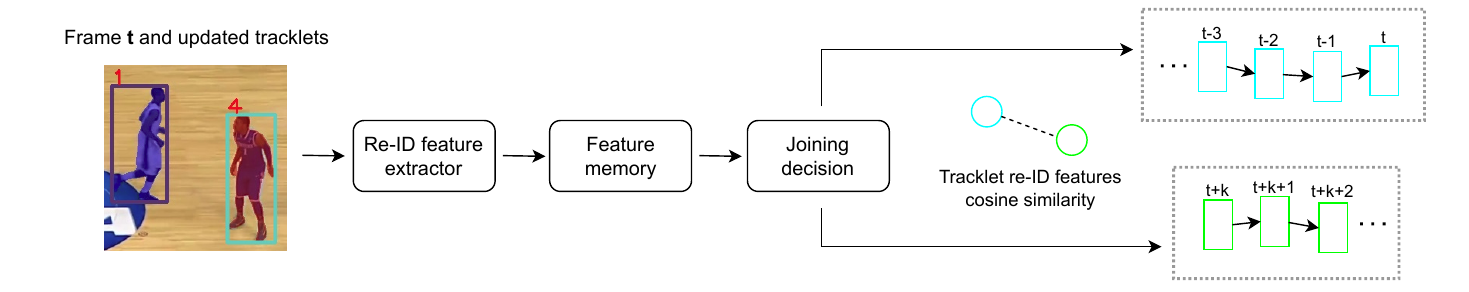}\\
(b) \\
\caption{
Overview of the proposed McByte++ online tracking pipeline.
(a) \textbf{Short-term tracking.}
\textbf{Top:} Tracklet bounding boxes from the previous frame $t\!-\!1$ are treated as the tracker state and propagated to frame $t$ using a Kalman filter to generate candidate tracklet positions.
\textbf{Middle:} 
Global camera motion is estimated and used to conditionally compensate the predicted tracklet positions, while the current frame $t$ is processed by the object detector.
An initial association cost matrix is computed based on the IoU between compensated tracklet candidates and detection bounding boxes.
\textbf{Bottom:} A mask temporal propagation module produces object masks at frame $t$, which are used as an auxiliary observation to enrich the association cost matrix under regulated conditions. The final cost matrix is solved using the Hungarian matching algorithm, yielding updated tracklets at frame $t$.
(b) \textbf{Long-term tracking.}
During short-term tracking, appearance-based re-identification features are extracted from tracked subjects using crops of their bounding boxes and stored in a feature memory. When a new tracklet is initialized at a later time $t\!+\!k$, its appearance representation is compared against those of previously terminated tracklets. Based on feature similarity, the new tracklet can be reconnected to and extend a corresponding long-term identity.
}
\label{fig:diagram_tracking_pipeline}
\end{figure*}

\section{Related Work}
\label{sec:rel_work}

\noindent\textbf{Multi-object tracking in sports.}
Multi-object tracking (MOT) is a key enabler for sports video analytics, supporting player localization, trajectory analysis, and tactical understanding. Sports datasets such as SportsMOT~\cite{sportsmot_ref} and SoccerNet-Tracking~\cite{soccernet-tracking2022_ref} highlight challenges that are less pronounced in generic pedestrian tracking, including fast and abrupt motion, severe occlusions, camera panning and zooming, and frequent entry and exit of tracked subjects. These properties place strong demands on robustness, efficiency, and identity consistency in tracking systems.

\noindent\textbf{Tracking-by-detection approaches.}
Tracking-by-detection remains a dominant paradigm for MOT due to its modularity and deployability. In this setting, objects are detected independently in each frame and associated over time using motion and geometric cues, typically within a Kalman-filter-based framework~\cite{kf_ref} and bipartite matching~\cite{hungarianalg_ref}. ByteTrack~\cite{bt_ref} and its extensions (e.g., OC-SORT~\cite{ocsort_ref}, Deep OC-SORT~\cite{deepocsort_ref}, StrongSORT~\cite{strongsort_ref}, HybridSORT~\cite{hybridsort_ref}) demonstrate that carefully designed association strategies can achieve strong performance without end-to-end training. However, many such methods rely on dataset- or sequence-specific hyper-parameter tuning and primarily address short-term association, limiting their robustness in sports scenarios with extended occlusions and re-entries.

\noindent\textbf{Recent learning-based MOT frameworks.}
Recent MOT research has increasingly explored learned temporal representations, spatial reasoning, and adaptive association strategies. Representative directions include synchronized sequence modeling~\cite{samba_ref}, transformer-based temporal feature fusion~\cite{gated_temporal_fusion_ref}, hypergraph-based spatial-temporal reasoning~\cite{hypergraph_mot_ref}, and test-time adaptation~\cite{tcei_ref}. These works illustrate the growing diversity of learned temporal and spatial reasoning strategies for robust multi-object tracking. However, they are generally evaluated under different detector configurations, adaptation mechanisms, training procedures, or experimental settings from the training-free framework considered in this work, making direct quantitative comparison less straightforward. In contrast, McByte++ adopts a fundamentally different design philosophy based on training-free tracking-by-detection, where robust long-term association is achieved through the selective integration of complementary tracking cues rather than learned association models.

\noindent\textbf{Mask temporal propagation and mask-based tracking.}
Temporal mask propagation models developed for video object segmentation (VOS) provide strong priors on object shape and temporal continuity. XMem~\cite{xmem_ref} introduced a memory-based framework for long-term mask propagation, while Cutie~\cite{cutie_ref} further improved object–background separation through enhanced memory usage. Image segmentation models such as SAM~\cite{sam_ref} generate high-quality single-frame masks but do not directly address temporal consistency. Several tracking systems attempt to build upon these components: DEVA~\cite{deva_ref} combines mask propagation with box tracking, Grounded SAM2~\cite{sam2_ref,gr_dino_ref} integrates grounding-based detection with SAM-style masks, and MASA~\cite{masa_ref} uses SAM-derived features for detection matching. Despite their strong segmentation capabilities, these approaches often lack robust tracklet management and struggle with missed detections, long occlusions, and frequent identity interruptions - conditions~\cite{MCBYTE_REF} common in broadcast sports footage.

\noindent\textbf{Re-identification in multi-object tracking.}
Appearance-based re-identification (re-ID)~\cite{osnet_ref,market1501_ref} has long been used to reduce identity switches and improve robustness under occlusion. Classical systems such as DeepSORT~\cite{deepsort_ref} integrate deep appearance embeddings into SORT-style~\cite{sort_ref} tracking, while more recent methods, including BoT-SORT~\cite{botsort_ref}, StrongSORT~\cite{strongsort_ref}, and Deep OC-SORT~\cite{deepocsort_ref}, combine motion cues with learned appearance features and, in some cases, camera motion compensation~\cite{strongsort_ref,orb_ref}. In most of these approaches, re-ID features are used at every association step to refine frame-level matching.

However, re-ID in MOT is not universally beneficial. Appearance embeddings extracted from partially occluded, motion-blurred, or low-resolution subjects can be unreliable, and aggressive use of re-ID may amplify association errors in crowded scenes. This issue is particularly relevant in sports, where players frequently overlap and visual conditions degrade rapidly. Recent analysis~\cite{reid_no_help_ref} systematically studies these effects and shows that re-ID often provides limited gains or even degrades performance when applied indiscriminately during short-term association.

\noindent\textbf{Trajectory forecasting for long-term tracking.}
Trajectory forecasting has emerged as a complementary direction for improving long-term multi-object tracking by exploiting motion histories to predict plausible future object locations under occlusion or temporary absence. Methods such as Quo Vadis~\cite{quovadis_ref} leverage motion prediction to support long-term identity continuity, while sports-oriented approaches including SportMamba~\cite{sportmamba_ref} explicitly model non-linear player motion. These methods are complementary to McByte++. Whereas they primarily improve association through learned motion prediction, our work focuses on training-free online identity reconnection through the selective integration of masks, re-identification, and camera motion compensation.

\noindent\textbf{Long-term identity association and post-processing.}
To improve long-term identity consistency, several works address tracking beyond short-term association by performing global or post-processing tracklet linking. Graph-based approaches such as SUSHI~\cite{sushi_ref} formulate tracking as a hierarchical association problem over extended temporal windows, but rely on offline optimization and introduce substantial computational overhead, making them challenging or even infeasible to apply to long, crowded, or high-resolution sports videos. Post-processing methods such as Global Tracklet Association~\cite{gta_link_ref} refine tracker outputs by first splitting tracklets to mitigate possible identity switches and subsequently linking them to recover long-term identities. While effective at improving identity continuity, these approaches operate as separate refinement stages rather than integrated online trackers, adding complexity and limiting their suitability for efficient online sports analytics.

\noindent\textbf{Positioning of McByte++.}
McByte++ builds upon the training-free tracking-by-detection framework introduced in McByte~\cite{MCBYTE_REF}, extending it from short-term mask-guided association toward efficient long-term identity preservation through the selective integration of complementary tracking cues.
In contrast to prior approaches that incorporate re-identification during every frame-level association stage, McByte++ decouples short-term association from long-term identity recovery by invoking re-identification only when new tracklets are initialized, allowing them to be connected to previously terminated identities after extended absences. This design enables long-term identity reconnection without relying on offline post-processing or dataset-specific training. McByte++ uses a lightweight, pretrained OSNet~\cite{osnet_ref} re-ID backbone, preserving efficiency and deployability while improving identity consistency in realistic sports scenarios. 

Several concepts employed in McByte++, including long-term memory, appearance representation, deformation-aware tracking, multi-scale processing, and motion modeling, have also been investigated in the single-object tracking literature~\cite{facing_occ_ref,intrinsic_rep_ref,deformation_slack_ref,split_frequency_ref,distance_constraint_ref}. While our work focuses on training-free online multi-object tracking in sports, these studies provide complementary perspectives on robust long-term visual tracking.

%TC:ignore
\section{Proposed method}
\label{sec:method}

\subsection{Preliminaries}
\label{sec:method_preliminaries}

Tracking-by-detection methods~\cite{bt_ref,ocsort_ref,deepocsort_ref,cbiou_ref,strongsort_ref,hybridsort_ref} associate detection bounding boxes of the same objects across frames to form \emph{tracklets}. At each time step, detections in the current frame are matched to existing tracklets from previous frames using association cues such as spatial proximity, motion consistency, and geometric overlap. These cues are combined to build a cost matrix that reflects the likelihood of each tracklet-detection pairing. The association problem is then formulated as a bipartite matching task and solved using the Hungarian algorithm~\cite{hungarianalg_ref}, with matches exceeding a predefined cost threshold being rejected. Matched detections extend existing tracklets, while unmatched detections may initialize new tracklets and tracklets unmatched for an extended period are terminated.

Similarly as McByte~\cite{MCBYTE_REF}, McByte++ builds upon ByteTrack~\cite{bt_ref} as its baseline tracking-by-detection framework. ByteTrack separates detections into high- and low-confidence sets and handles them in successive association stages. Association relies primarily on the intersection-over-union (IoU) metric between predicted tracklet bounding boxes and observed detections. Tracklet states are propagated using a linear motion model implemented via a Kalman filter~\cite{kf_ref}, and the IoU-based cost matrix is constructed as $1-\mathrm{IoU}$. The reader is referred to the original ByteTrack work for full implementation details.

The goal of McByte++ is to develop a robust and efficient tracking system for sports videos that operates \emph{without any training or dataset-specific tuning}. To this end, McByte++ augments the baseline tracking-by-detection pipeline with additional cues and mechanisms that improve robustness under challenging sports conditions, 
while maintaining reasonable runtime. 

Rather than continuously applying all available association cues throughout tracking, McByte++ adopts a selective integration strategy in which each cue is used only when it is expected to provide complementary information. Temporally propagated masks assist short-term association in ambiguous or isolated situations, camera motion compensation is applied only when reliable global motion can be estimated, and re-identification is reserved for reconnecting newly initialized tracklets with previously terminated long-term identities. This design preserves the efficiency of the underlying tracking-by-detection framework while extending its capability toward robust long-term tracking.

The individual components of McByte++ are described in the following subsections. We first present the mask-guided association inherited from McByte~\cite{MCBYTE_REF}, followed by the proposed long-term re-identification mechanism (Sec.~\ref{sec:method_reid}) and the improved conditional camera motion compensation strategy (Sec.~\ref{sec:method_mcbyte_with_cmc}).

An overview of the complete McByte++ tracking pipeline, including mask-based association, long-term re-identification, and camera motion compensation, is shown in Fig.~\ref{fig:diagram_tracking_pipeline}.

\subsection{Mask creation and handling}
\label{sec:method_mask_management}

To enrich the association process beyond bounding box geometry, McByte++ leverages temporally propagated segmentation masks as an additional cue. Each active tracklet is associated with a corresponding object mask that is propagated across frames and updated online. In contrast to the original McByte, which relied on a heavier mask propagation model, McByte++ adopts EdgeTAM~\cite{edgetam_ref} as its mask propagation module.

EdgeTAM is a lightweight mask propagation model designed for efficient temporal mask tracking. Its use significantly reduces computational overhead compared to earlier propagation approaches, making it better suited for 
sports analytics. In McByte++, EdgeTAM is applied in an end-to-end manner: once a tracklet is initialized, its associated mask is propagated forward in time and updated synchronously with the tracklet state.

The propagated masks are maintained in alignment with the lifespan of tracklets and are updated at each frame as part of the tracking loop. These masks are not used as standalone tracking outputs; instead, they serve as an auxiliary signal to support data association, particularly in challenging scenarios involving close interactions, partial occlusions, or abrupt motion. By combining lightweight mask propagation with a regulated integration strategy described in Sec.~\ref{sec:method_mask_use}, McByte++ is able to exploit spatial object extent information while preserving efficiency.

Overall, replacing the heavier mask propagation component used in the original McByte with EdgeTAM substantially reduces runtime cost. When combined with a controlled use of mask information in the association process, this design allows McByte++ to retain the benefits of mask-based cues while contributing to the improved speed and practicality of the overall system.

\begin{figure}
\centering

\includegraphics[width=6cm]{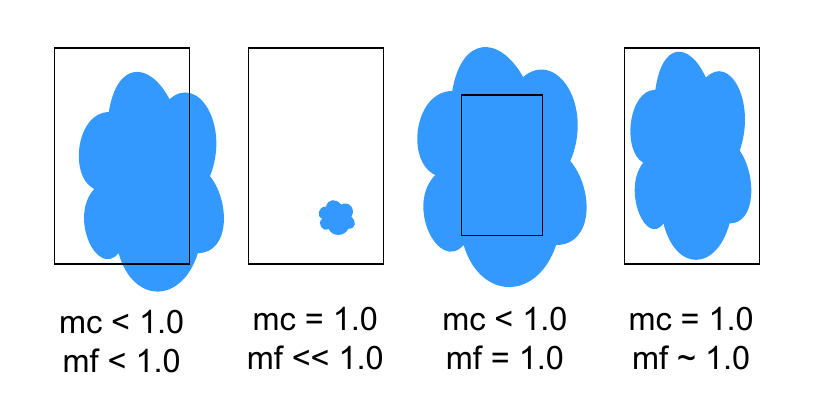} \\
(a) \\
\includegraphics[width=8cm] {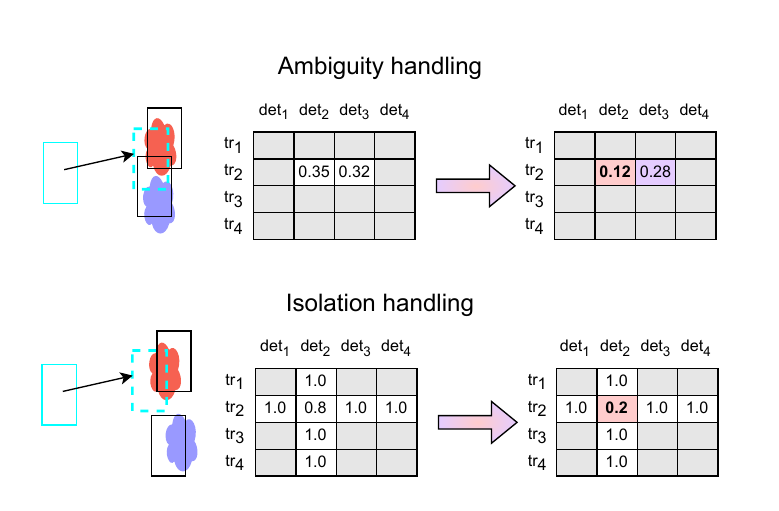} \\ 
(b) \\
\caption{
(a) Examples illustrating how the bounding box coverage ($mc$) and mask fill ratio ($mf$) vary depending on the relative position of a temporally propagated mask (blue) and a detection bounding box. The most favorable case for mask-based guidance occurs when both $mc$ and $mf$ are close to $1$. 
(b) Illustration of ambiguity and isolation cases in the association process, where temporally propagated masks can provide additional guidance. Ambiguity arises when IoU-based costs are similarly low for multiple entries in a row or column of the cost matrix, while isolation occurs when relevant IoU-based costs are too high to allow association despite the absence of ambiguity.
}
\label{fig:ambig_isol_handling_and_mm1_mm2}
\end{figure}

\subsection{Regulated use of the mask}
\label{sec:method_mask_use}

Although temporally propagated masks provide valuable spatial information, they can occasionally be inaccurate due to drift, partial occlusions, or rapid motion. Therefore, McByte++ integrates mask information into the association process in a \emph{regulated} manner, ensuring that mask cues influence matching decisions only when they are likely to be reliable.

During data association, mask-based guidance is applied only in two specific situations: \emph{ambiguity} and \emph{isolation}. Ambiguity occurs when a detection can plausibly match multiple tracklets, or vice versa, typically due to strong overlap between bounding boxes leading to similar IoU scores. Isolation refers to cases where a tracklet and a detection are spatially distant and their IoU-based cost exceeds the matching threshold, despite potentially corresponding to the same object. Such situations commonly arise under motion blur or abrupt camera movement, where short-term geometric cues alone are insufficient.

For each candidate tracklet-detection pair in these cases, McByte++ applies a set of gating conditions before incorporating mask information. Specifically, the following conditions must hold:
\begin{enumerate}
    \item The temporally propagated mask associated with the tracklet is visible in the current frame.
    \item The mask sufficiently overlaps the detection bounding box, as measured by the mask fill ratio.
    \item The detection bounding box sufficiently covers the propagated mask, as measured by the bounding box coverage.
\end{enumerate}

To formalize these checks, we define two ratios based on pixel counts. Let $mask(tracklet_i)$ denote the propagated mask associated with tracklet $i$, and let $bbox_j$ denote detection $j$. The bounding box coverage of the mask is defined as:
\begin{equation}
mc^{i,j} = \frac{|mask(tracklet_i) \cap bbox_j|}{|mask(tracklet_i)|},
\end{equation}
and the mask fill ratio of the bounding box is defined as:
\begin{equation}
mf^{i,j} = \frac{|mask(tracklet_i) \cap bbox_j|}{|bbox_j|},
\end{equation}
where $|\cdot|$ denotes set cardinality and both $mc^{i,j}$ and $mf^{i,j}$ take values in $[0,1]$. In Fig.~\ref{fig:ambig_isol_handling_and_mm1_mm2}(a), we illustrate how $mc$ and $mf$ vary depending on the relative position of the temporally propagated mask and the detection bounding box. In all experiments, the minimum mask fill ratio threshold is set to $mf = 0.05$, while the minimum mask coverage threshold is set to $mc = 0.90$. These values are kept fixed across all datasets and experiments without dataset-specific adjustment. Their consistent use across multiple sports benchmarks, camera viewpoints, and scene conditions demonstrates that the proposed operating points generalize well in practice while remaining consistent with the training-free philosophy of McByte++.

Only when all conditions above are satisfied is the IoU-based association cost updated using the mask cue:
\begin{equation}
  costs^{i,j} =
  \begin{cases}
    costs^{i,j}_{\mathrm{IoU}} - mf^{i,j}, & \text{if cond. (1)-(3) satisfied}, %\\
    \\[1ex]
    costs^{i,j}_{\mathrm{IoU}}, & \text{otherwise}.
  \end{cases}
  \label{eq:cost_matrix_update}
\end{equation}
Here, $costs^{i,j}_{\mathrm{IoU}}$ denotes the original IoU-based cost between tracklet $i$ and detection $j$, and $costs^{i,j}$ denotes the final association cost. Fig.~\ref{fig:ambig_isol_handling_and_mm1_mm2}(b) shows how the temporally propagated mask influences the cost matrix and guides the association toward the correct tracklet-detection match.

In this formulation, $mc^{i,j}$ is used strictly as a gating criterion rather than as a cost modifier, since multiple masks may fully lie within the same bounding box and yield identical coverage values. In contrast, $mf^{i,j}$ provides a more discriminative measure of how well a detection spatially aligns with a propagated mask and is therefore used to influence the association cost.

By regulating when and how mask information is incorporated, McByte++ avoids over-reliance on potentially noisy mask predictions. When propagated masks become inaccurate due to drift, severe occlusions, or rapid motion, the proposed gating strategy prevents unreliable mask cues from influencing the association process. Temporally propagated masks are integrated as an auxiliary association cue only under controlled conditions, guiding the Hungarian matching process in ambiguous or isolated cases without altering the underlying matching formulation. This design allows lightweight mask propagation to effectively assist association in challenging scenarios, while preserving robustness, efficiency, and the core structure of the baseline tracking pipeline.

\subsection{Online Re-Identification for Long-Term Tracking}
\label{sec:method_reid}

Sports video analytics frequently requires maintaining identity consistency over extended temporal gaps caused by occlusions, camera motion, or subjects leaving and re-entering the field of view. While short-term association in tracking-by-detection pipelines can be handled reliably using motion, geometry, and mask-based cues, these signals alone are insufficient for long-term identity recovery. McByte++ addresses this challenge through a selective, online re-identification (re-ID) mechanism designed specifically for long-term tracking in sports.

\noindent\textbf{Design principle.}
A key design choice in McByte++ is to decouple re-identification from short-term association. Appearance-based cues are not used during frame-to-frame matching or integrated into the primary assignment cost. Instead, re-ID is invoked selectively to reconnect newly initialized tracklets with previously terminated long-term identities. This avoids known failure modes of re-ID under partial occlusion, motion blur, and crowded interactions, while preserving the robustness of motion- and mask-based short-term tracking.

\noindent\textbf{Appearance representation.}
Appearance descriptors are extracted using a pretrained OSNet re-identification backbone~\cite{osnet_ref}, which provides a lightweight and efficient representation suitable for online use. The pretrained model is adopted directly without any fine-tuning on the target datasets, following common practice in tracking systems that employ off-the-shelf re-ID models~\cite{gta_link_ref}.

For each reliable detection associated with a tracklet, an appearance feature vector $\mathbf{f}_t \in \mathbb{R}^{d}$ is extracted from the corresponding bounding box and $\ell_2$-normalized:
\begin{equation}
\tilde{\mathbf{f}}_t = \frac{\mathbf{f}_t}{\lVert \mathbf{f}_t \rVert_2}.
\end{equation}

\noindent\textbf{Temporal aggregation of appearance features.} Single-frame appearance descriptors are often noisy in sports videos due to fast motion and partial occlusions. To improve robustness, McByte++ aggregates appearance information over time at the tracklet level. For a given tracklet with a set of normalized descriptors $\{\tilde{\mathbf{f}}_1, \dots, \tilde{\mathbf{f}}_N\}$, a tracklet-level representation is computed as:
\begin{equation}
\mathbf{f}_{\mathrm{avg}} = \frac{1}{N} \sum_{i=1}^{N} \tilde{\mathbf{f}}_i
\end{equation}
\begin{equation}
\tilde{\mathbf{f}}_{\mathrm{avg}} =
\frac{\mathbf{f}_{\mathrm{avg}}}{\lVert \mathbf{f}_{\mathrm{avg}} \rVert_2}
\end{equation}
This aggregation stabilizes the appearance representation and mitigates the influence of unreliable individual observations.

\noindent\textbf{Long-term identity memory.}
When a tracklet is terminated (e.g., due to extended occlusion or exiting the scene), its aggregated descriptor $\tilde{\mathbf{f}}_{\mathrm{avg}}$ is stored in a long-term identity memory. Each entry in this memory represents a previously observed long-term identity and contains only minimal information required for future reconnection. The memory is maintained online and updated dynamically during tracking.

\noindent\textbf{Online re-identification and similarity computation.}
For each newly initialized tracklet, McByte++ attempts to reconnect it to a stored long-term identity using appearance similarity. Given the aggregated descriptor of a provisional tracklet $\tilde{\mathbf{f}}^{(p)}$ and a stored identity descriptor $\tilde{\mathbf{f}}^{(m)}$, similarity is computed using cosine similarity:
\begin{equation}
\mathrm{sim}\!\left(\tilde{\mathbf{f}}^{(p)}, \tilde{\mathbf{f}}^{(m)}\right)
=
\tilde{\mathbf{f}}^{(p)} \cdot \tilde{\mathbf{f}}^{(m)}
\end{equation}
Since both descriptors are $\ell_2$-normalized, this corresponds to the cosine of the angle between the feature vectors.

A long-term identity is transferred (i.e. a reconnection is accepted) only if the correspondence satisfies a mutual best-match criterion. Specifically, a newly initialized tracklet must identify a terminated long-term identity as its most similar candidate, while that terminated identity must also identify the same new tracklet as its highest-similarity match among all candidate tracklets. Identity transfer is performed only when both conditions are satisfied and the similarity exceeds the predefined threshold.

\noindent\textbf{Identity assignment and finalization.}
If a valid match is found, the provisional tracklet inherits the corresponding long-term identity, and the identity is considered finalized. The associated memory entry is then removed to prevent duplicate assignments. If no reconnection occurs within a short temporal window, the tracklet is confirmed as a new long-term identity. This delayed finalization avoids premature identity assignment and reduces the risk of later conflicts.

By activating re-identification only for long-term identity recovery and aggregating appearance information over time, McByte++ leverages re-ID where it is most reliable while avoiding its pitfalls in short-term association. Unlike approaches that rely on appearance cues throughout frame-to-frame association, McByte++ uses re-identification only for reconnecting newly initialized tracklets with previously terminated long-term identities. This selective strategy substantially reduces the influence of occasional appearance ambiguities between visually similar players while preserving the benefits of appearance-based long-term identity recovery. Nevertheless, very similar teammates may still occasionally remain challenging, particularly under significant appearance changes or limited visual evidence. This design enables online long-term tracking without additional training, offline processing, or heavy computational overhead.

\subsection{Conditional Camera Motion Compensation}
\label{sec:method_mcbyte_with_cmc}

Sports broadcast videos frequently exhibit strong global camera motion due to panning, zooming, and viewpoint changes. If not accounted for, such motion can significantly degrade prediction accuracy and association quality in tracking-by-detection pipelines. As in the original McByte framework, McByte++ incorporates camera motion compensation to align track predictions with global frame-to-frame motion. However, McByte++ revises the way camera motion compensation is applied, addressing key robustness and efficiency limitations observed in the original formulation.

\noindent\textbf{Motivation and limitations of unconditional compensation.}
In the original McByte, camera motion compensation was applied unconditionally whenever enabled. While effective under favorable conditions, this strategy could be harmful when motion estimation was unreliable, for instance in scenes with weak texture, heavy motion blur, or largely homogeneous backgrounds. In such cases, inaccurate motion estimates could lead to distorted bounding boxes, unstable Kalman predictions, and severe degradation of tracking performance. This observation motivates a more controlled use of camera motion compensation.

\noindent\textbf{\noindent\textbf{Conditional application of camera motion compensation.}
}
McByte++ replaces unconditional camera motion correction with a conditional application strategy. Camera motion compensation is applied only when the estimated global motion satisfies a set of reliability criteria designed to prevent degenerate transformations. These criteria ensure that the estimated affine transform is physically plausible and does not induce excessive scaling or translation that could collapse or explode bounding boxes. When the estimated motion does not meet these conditions, the tracker proceeds without applying camera motion correction, thereby avoiding harmful updates.

This reliability-aware design preserves the benefits of camera motion compensation when informative motion cues are present, while preventing failure cases when motion estimation is unreliable.

\noindent\textbf{\noindent\textbf{Motion estimation with spatial downscaling.}}
In both McByte~\cite{MCBYTE_REF} and McByte++, global camera motion between consecutive frames is modeled as an affine transformation on tracked object positions (their bounding boxes) and estimated using ORB-based feature matching~\cite{orb_ref}. To improve robustness and efficiency in McByte++, frames are spatially downscaled prior to feature extraction and matching. Downscaling reduces sensitivity to noise and motion blur, stabilizes feature correspondences, and significantly lowers the computational cost of motion estimation.

While the original McByte used a fixed downscaling factor of 2, McByte++ explores larger downscaling factors and adopts a factor of 4 as a more effective trade-off between accuracy and efficiency. This choice improves the stability of motion estimation in challenging sports scenes and contributes to the overall runtime speed-up of McByte++. A detailed analysis of this trade-off is provided in the experimental evaluation and discussion.

\noindent\textbf{\noindent\textbf{Temporal consistency and robustness.}}
To further stabilize camera motion compensation over time, McByte++ maintains the most recent reliable motion estimate. If a newly computed estimate is deemed unreliable, the system avoids updating the motion correction, preventing abrupt frame-to-frame inconsistencies. This temporal consistency mechanism improves robustness in sequences with intermittent feature degradation or rapid camera changes.

\noindent\textbf{\noindent\textbf{Integration and efficiency considerations.}}
Camera motion compensation is applied after Kalman filter prediction and before data association, ensuring that predicted track states are aligned with global camera motion prior to matching. Importantly, conditional application and increased downscaling reduce unnecessary computation, making camera motion compensation both safer and more efficient. Alongside other system-level improvements, this contributes to the improved runtime performance of McByte++ compared to the original McByte, while maintaining or improving tracking accuracy.

%TC:endignore

\section{Experiments and discussion}

\subsection{Implementation details}

McByte++ is built upon the ByteTrack~\cite{bt_ref} tracking-by-detection framework. For object detections, we follow the baseline setup and use YOLOX~\cite{yolox_ref} detectors pre-trained on the respective datasets, unless stated otherwise. We do not train or fine-tune detectors within this work. To ensure fair comparison, we use the same detections as the baseline and other compared tracking methods whenever possible.

Detections are divided into high- and low-confidence groups following ByteTrack. Unlike some approaches that tune confidence thresholds per sequence, we employ a fixed confidence threshold of $0.6$ across all datasets and sequences, maintaining a training-free and tuning-free configuration.

For mask propagation, McByte++ uses EdgeTAM~\cite{edgetam_ref} as described in Sec.~\ref{sec:method_mask_management}. In contrast to the original McByte~\cite{MCBYTE_REF}, which relied on a heavier propagation model, EdgeTAM provides a lightweight alternative with substantially lower computational overhead while maintaining sufficient mask quality for association guidance. Mask-based association follows the regulated strategy defined in Sec.~\ref{sec:method_mask_use}.

Camera motion compensation (CMC) is modeled as an affine transformation estimated via ORB-based feature matching (Sec.~\ref{sec:method_mcbyte_with_cmc}). We evaluate both unconditional (original, as in McByte) and conditional CMC variants, and explore spatial downscaling factors $ds \in \{2,4,6\}$ for motion estimation. Unless stated otherwise, the selected default configuration uses conditional CMC with $ds=4$, as motivated by the ablation study.

For the re-identification component (Sec.~\ref{sec:method_reid}), we employ a pretrained OSNet backbone. Appearance features are extracted online from bounding box crops, $\ell_2$-normalized, temporally aggregated, and compared using cosine similarity. We evaluate similarity thresholds $\{0.6, 0.7, 0.8, 0.9\}$ and select $0.8$ as the default configuration based on ablation results. No re-ID model training or dataset-specific tuning is performed.

All experiments are conducted in a training-free manner. The same parameter settings are used across all sequences and datasets, demonstrating the generality and practical applicability of McByte++ for sports tracking.

\subsection{Datasets and metrics}
\label{sec:experiments_discussion}

We evaluate McByte++ on sports-oriented multi-object tracking benchmarks, including SportsMOT~\cite{sportsmot_ref}, SoccerNet-tracking 2022~\cite{soccernet-tracking2022_ref}, and SoccerNet-tracking Challenge 2023. All datasets operate at 25 FPS.

SportsMOT~\cite{sportsmot_ref} contains basketball, volleyball, and soccer sequences captured from diverse indoor and outdoor court views. The dataset features rapid player motion, frequent occlusions, and strong camera movement. We use YOLOX detections pre-trained on SportsMOT following community standards for fair comparison.

SoccerNet-tracking 2022~\cite{soccernet-tracking2022_ref} consists of soccer match videos with visually similar players within teams and continuous camera motion. For the test split, oracle detections are provided, allowing evaluation of pure tracking performance.

SoccerNet-tracking Challenge 2023 extends the benchmark by new sequences, yet without provided detections. For this split, we use YOLOX trained on SportsMOT to generate detections. Due to the unexpected unavailability of the evaluation server during the preparation of this work, comparisons on this split are limited to McByte and McByte++ variants.

Across all datasets, we use identical detections for McByte++ and other tracking methods whenever possible to ensure fair comparison. No detector retraining is performed.

We report three standard MOT metrics: HOTA~\cite{hota_ref}, IDF1~\cite{idf1_ref}, and MOTA~\cite{mota_ref}. HOTA jointly evaluates detection, localization, and association quality, while IDF1 measures identity preservation and long-term consistency. MOTA primarily reflects detection accuracy and is included for completeness. Since McByte++ focuses on improving identity consistency and long-term tracking, we emphasize HOTA and IDF1 in the analysis, while interpreting MOTA cautiously when oracle detections are provided.

\noindent\textbf{Data Availability.} The datasets analysed during the current study are available in the following repositories: SoccerNet-tracking~\cite{soccernet-tracking2022_ref, soccernet-tracking2023_ref} at \url{https://github.com/SoccerNet/sn-tracking} and SportsMOT~\cite{sportsmot_ref} at \url{https://github.com/MCG-NJU/SportsMOT}.

\begin{table}
  \centering
  {\small{
  \scalebox{0.9}{
  \begin{tabular}{lcccc}
    \toprule
    Method & HOTA$\uparrow$ & IDF1$\uparrow$ & MOTA$\uparrow$ & FPS$\uparrow$ \\
    \midrule
    \multicolumn{5}{c}{Unconditional camera motion compensation -- no re-ID} \\
    % \midrule
    \rowcolor{lightyellow}
    McByte~\cite{MCBYTE_REF}       & 85.0 & 79.9 & 96.8 & 1.04 \\
    McByte++, ds=2 $\dagger$       & 84.6 & 79.5 & 97.0 & 6.49 \\
    McByte++, ds=4                 & 84.1 & 78.9 & 97.1 & 10.51 \\   
    McByte++, ds=6                 & 83.4 & 78.0 & 97.0 & 10.33 \\
    \midrule
    \multicolumn{5}{c}{Conditional camera motion compensation -- no re-ID} \\
    % \midrule
    McByte++, ds=2                 & 84.6 & 79.5 & 97.0 & 6.58 \\
    \rowcolor{lightgreen}
    McByte++, ds=4                 & 84.1 & 78.9 & 97.1 & 10.71 \\   
    McByte++, ds=6                 & 83.1 & 77.7 & 97.0 & 10.45 \\
    \midrule
    \multicolumn{5}{c}{With re-ID, online processing} \\
    % \midrule
    McByte++ re-ID sim\_th=0.6      & 85.7 & 82.9 & 97.1 & 8.70 \\
    McByte++ re-ID sim\_th=0.7      & 86.9 & 84.4 & 97.1 & 8.69 \\
    \rowcolor{lightblue}
    McByte++ re-ID sim\_th=0.8      & 87.5 & 84.5 & 97.1 & 8.69 \\
    McByte++ re-ID sim\_th=0.9      & 85.0 & 80.3 & 97.1 & 8.67 \\
    \midrule
    \multicolumn{5}{c}{With re-ID, offline post-processing} \\
    % \midrule
    \rowcolor{lightblue}
    McByte++ with GTA & 88.6 & 87.2 & 97.1 & 7.46 \\
    \bottomrule
  \end{tabular}
  }
  }}

  \caption{
   Ablation study on SoccerNet-tracking 2022 (test split) evaluating the impact of mask replacement, camera motion compensation (CMC) strategy, spatial downscaling factor ($ds$), and online re-identification. Results are reported using identical oracle detections. Runtime (FPS) improvements resulting from lightweight mask propagation and optimized CMC are explicitly reported. The original McByte is highlighted in yellow. The selected default McByte++ configuration (conditional CMC with $ds=4$) is highlighted in green. The best online re-ID configuration and the offline re-ID configuration are highlighted in blue.
   }
  \label{tab:cmc_not_fixed_and_fixed_soccernet_test}
\end{table}

\begin{table}
  \centering
  {\small{
  \scalebox{0.9}{
  \begin{tabular}{lcccc}
    \toprule
    Method & HOTA$\uparrow$ & IDF1$\uparrow$ & MOTA$\uparrow$ & FPS$\uparrow$ \\
    \midrule
    \multicolumn{5}{c}{Unconditional camera motion compensation -- no re-ID} \\
    % \midrule
    \rowcolor{lightyellow}
    McByte~\cite{MCBYTE_REF}       & 76.9 & 77.5 & 97.2 & 3.60 \\
    McByte++, ds=2 $\dagger$       & 75.9 & 76.3 & 96.9 & 11.88 \\
    McByte++, ds=4                 & 75.9 & 76.3 & 96.9 & 18.65 \\   
    McByte++, ds=6                 & 72.2 & 71.3 & 96.2 & 13.68 \\
    \midrule
    \multicolumn{5}{c}{Conditional camera motion compensation -- no re-ID} \\
    % \midrule
    McByte++, ds=2                 & 75.9 & 76.2 & 96.9 & 12.07 \\
    \rowcolor{lightgreen}
    McByte++, ds=4                 & 75.8 & 76.0 & 96.9 & 19.08 \\   
    McByte++, ds=6                 & 73.8 & 73.3 & 96.6 & 16.78 \\
    \midrule
    \multicolumn{5}{c}{With re-ID, online processing} \\ % buffered
    % \midrule
    McByte++ re-ID sim\_th=0.6      & 78.9 & 83.2 & 96.9 & 14.54 \\
    McByte++ re-ID sim\_th=0.7      & 79.5 & 83.7 & 96.9 & 14.53 \\
    \rowcolor{lightblue}
    McByte++ re-ID sim\_th=0.8      & 79.9 & 83.6 & 96.9 & 14.57 \\
    McByte++ re-ID sim\_th=0.9      & 76.8 & 77.6 & 96.9 & 14.57 \\
    \midrule
    \multicolumn{5}{c}{With re-ID, offline post-processing} \\
    % \midrule
    \rowcolor{lightblue}
    % \rowcolor{lightviolet}
    McByte++ with GTA & 81.5 & 86.0 & 96.8 & 12.49 \\
    \bottomrule
  \end{tabular}
  }
  }}
  \caption{
Ablation study on SportsMOT (test split) under identical detection settings. The table evaluates unconditional and conditional CMC, spatial downscaling factors ($ds$), and online re-identification thresholds. Runtime improvements resulting from lightweight mask propagation and optimized CMC are explicitly reported. Color highlights follow the same convention as in Table~\ref{tab:cmc_not_fixed_and_fixed_soccernet_test}. 
}
  \label{tab:cmc_not_fixed_and_fixed_sportsmot_test}
\end{table}

\subsection{Ablation study}

We perform extensive ablation studies on SoccerNet-tracking 2022 (test split) and SportsMOT (test split) to evaluate the contribution of individual components of McByte++, including camera motion compensation (CMC), spatial downscaling, and re-identification. All variants use EdgeTAM for mask propagation.

\paragraph{Impact of mask replacement and CMC downscaling (unconditional CMC).}

We first evaluate McByte++ without re-identification, using the same unconditional CMC strategy as in the original McByte and varying the spatial downscaling factor $ds \in \{2,4,6\}$. The original McByte results are shown for reference (highlighted in yellow in Tables~\ref{tab:cmc_not_fixed_and_fixed_soccernet_test} and~\ref{tab:cmc_not_fixed_and_fixed_sportsmot_test}). The variant with $\dagger$ denotes a configuration where only the mask propagation model is changed (from Cutie~\cite{cutie_ref} in McByte to EdgeTAM~\cite{edgetam_ref} in McByte++), showing its impact, while all the other settings remain the same between the two tracking algorithms.

On SoccerNet-tracking, replacing the original mask propagation module with EdgeTAM while keeping $ds=2$ yields comparable tracking performance (HOTA 84.6 vs. 85.0, IDF1 79.5 vs. 79.9), but dramatically improves runtime from 1.04 FPS to 6.49 FPS, representing more than a sixfold speed increase. Increasing $ds$ to 4 further boosts runtime to 10.51 FPS with only minor performance degradation (HOTA 84.1, IDF1 78.9). On SportsMOT, where runtime increases from 3.60 FPS (McByte) to 18.65 FPS (McByte++, $ds=4$), HOTA and IDF1 values drop by around 1\%.

These results demonstrate that replacing the heavy mask propagation module with EdgeTAM substantially reduces computational cost while preserving similar tracking accuracy.

\paragraph{Conditional camera motion compensation.}

We next evaluate the conditional CMC strategy described in Sec.~\ref{sec:method_mcbyte_with_cmc}. Compared to unconditional CMC, conditional compensation maintains similar HOTA and IDF1 values but further stabilizes performance and slightly improves runtime. On SoccerNet-tracking with $ds=4$, conditional CMC achieves 10.71 FPS compared to 10.51 FPS with unconditional CMC, while maintaining identical HOTA (84.1). On SportsMOT, conditional CMC with $ds=4$ reaches 19.08 FPS.

Across both datasets, $ds=4$ consistently provides the best trade-off between accuracy and efficiency. Therefore, we adopt conditional CMC with $ds=4$ (highlighted in green) as the default configuration for subsequent experiments.

\paragraph{Online re-identification.}

We then integrate the online re-identification mechanism described in Sec.~\ref{sec:method_reid}. On SoccerNet-tracking, adding re-ID significantly improves identity-related metrics. With similarity threshold $0.8$, HOTA increases from 84.1 to 87.5 (+3.4), and IDF1 increases from 78.9 to 84.5 (+5.6), while runtime remains at 8.69 FPS. Similar improvements are observed on SportsMOT, where HOTA improves from 75.8 to 79.9 (+4.1) and IDF1 from 76.0 to 83.6 (+7.6). MOTA remains largely unchanged, confirming that improvements stem primarily from enhanced identity preservation rather than detection differences.

The similarity threshold study shows that overly low thresholds (e.g., 0.6) introduce incorrect long-term associations by allowing visually similar yet distinct players to be merged into the same identity. Conversely, overly high thresholds (0.9) become too restrictive and may prevent correct reconnection of the same player after appearance changes caused by pose variation, viewpoint changes, or illumination differences. A threshold of 0.8 consistently provides the best overall balance between these two effects, achieving the highest HOTA and IDF1 across the evaluated datasets, while MOTA and runtime remain largely unaffected. Consequently, 0.8 is selected as a robust operating point and used in all subsequent experiments (highlighted in blue).

\begin{figure*}
\centering
\begin{tabular}{cc}
Visualizations without masks & Visualizations with masks\\
\includegraphics[height=4.4cm]{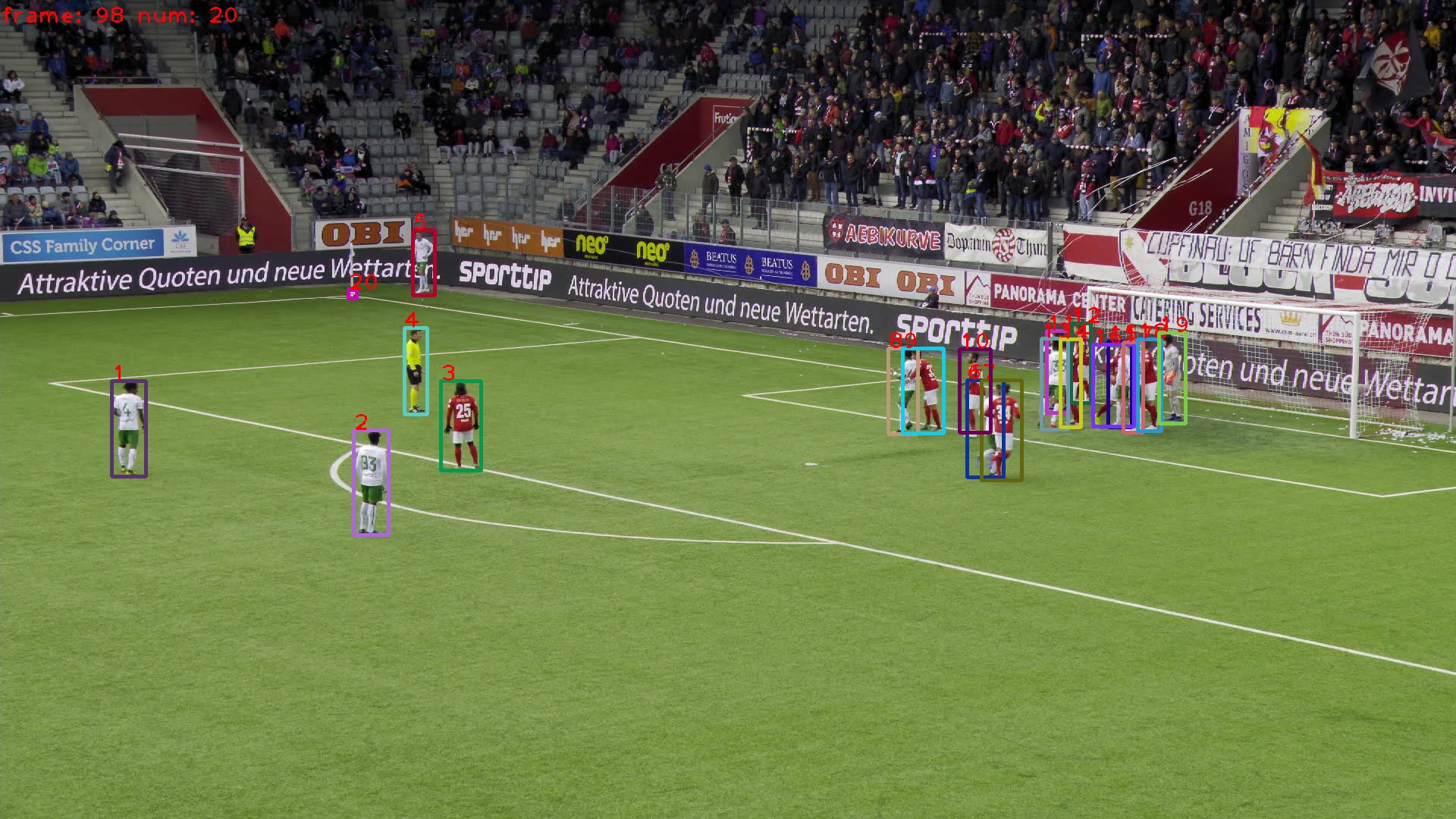}  &
\includegraphics[height=4.4cm]{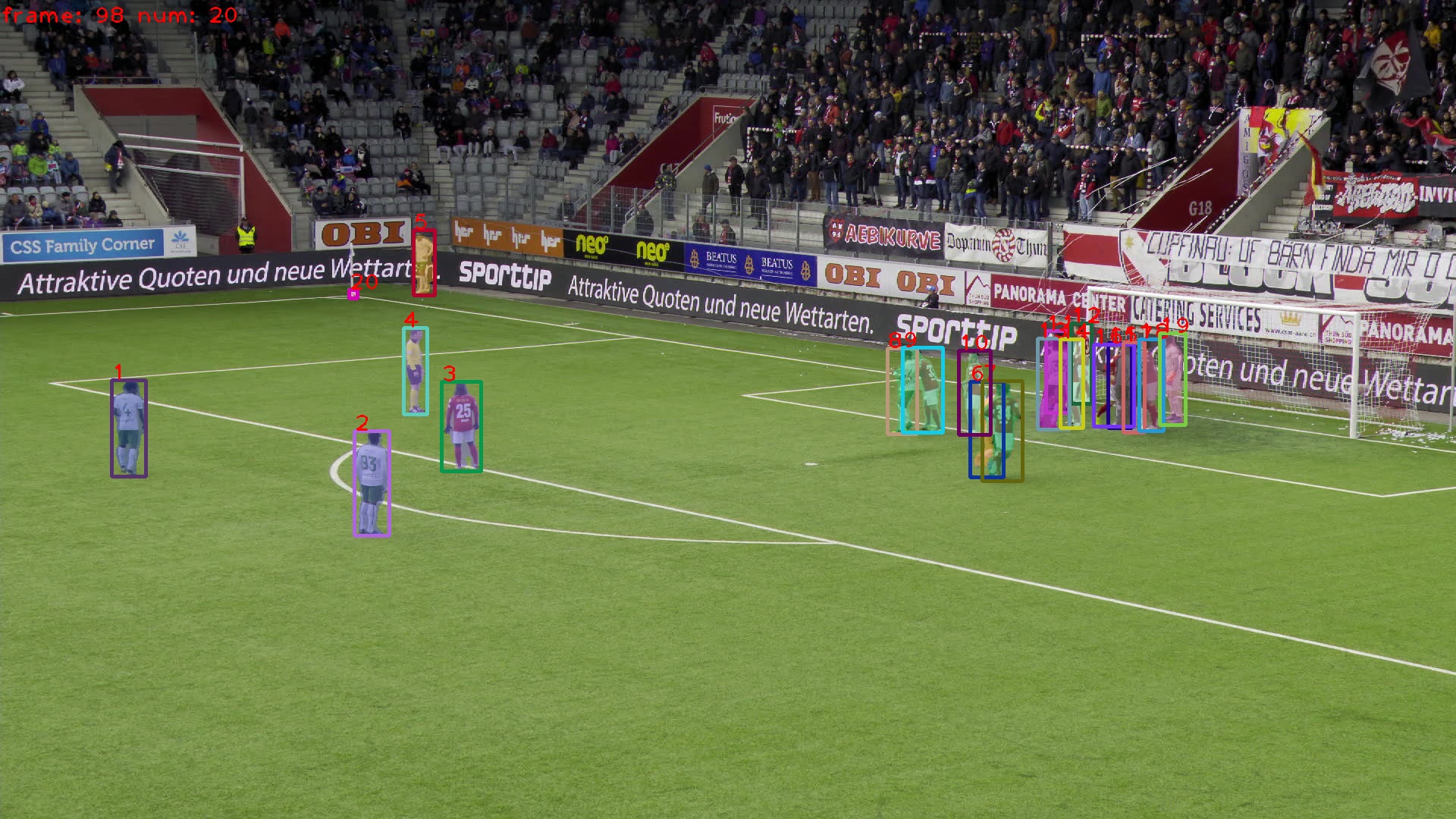}
\\
\multicolumn{2}{c}{Frame 98 - players with IDs 1 and 2 present.}
\\
\includegraphics[height=4.4cm]{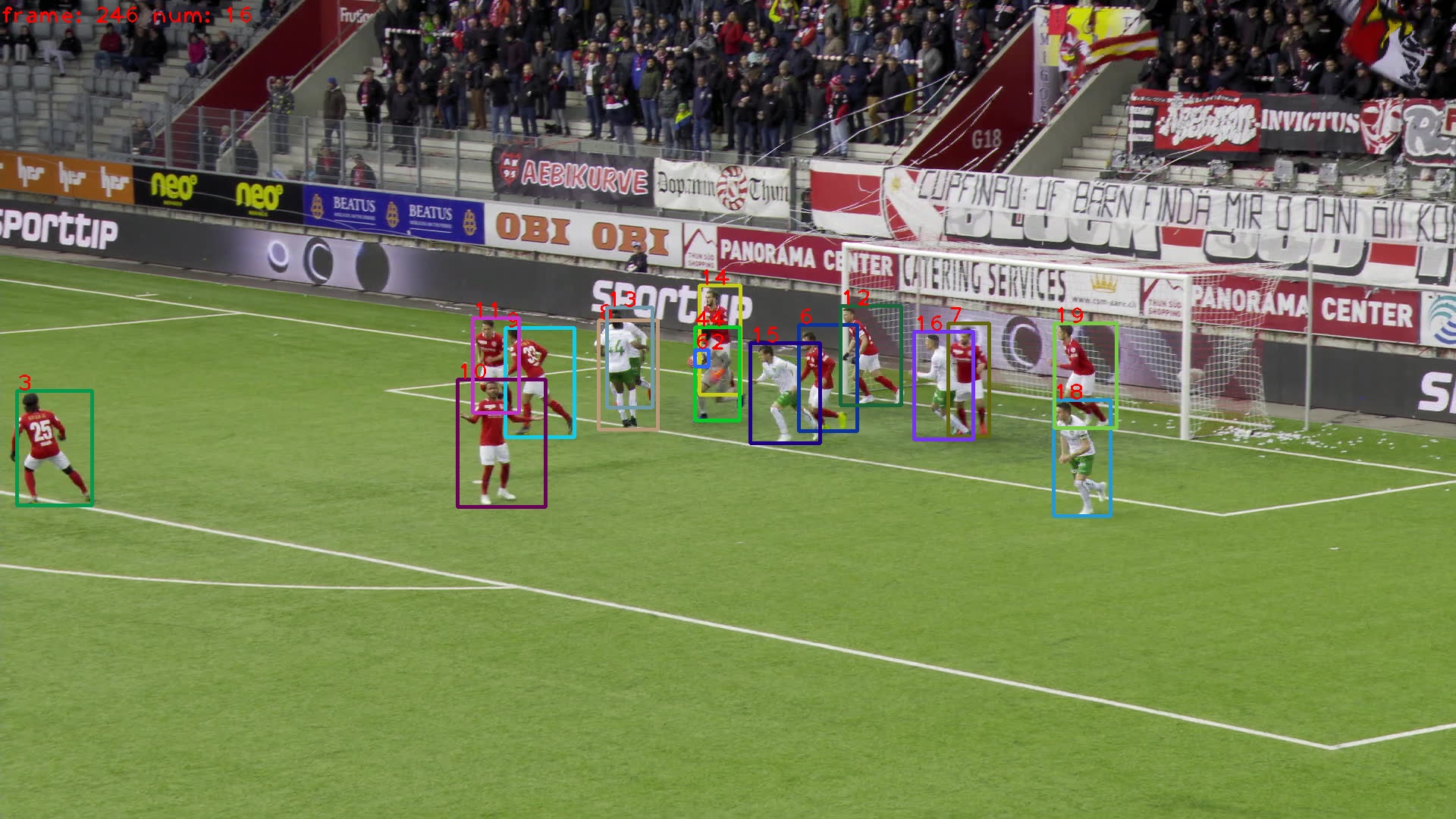}  &
\includegraphics[height=4.4cm]{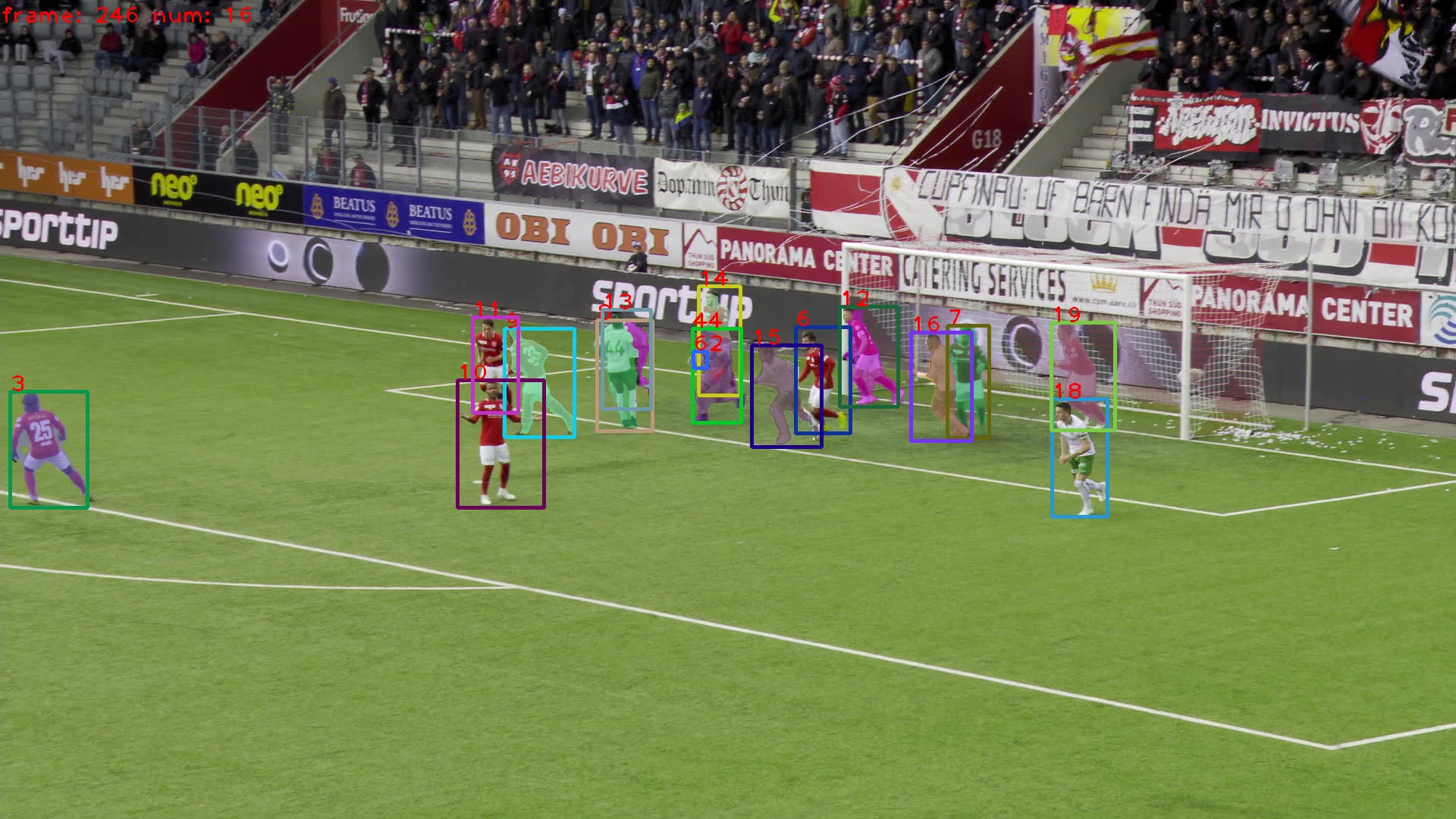}
\\
\multicolumn{2}{c}{Frame 246 - players with IDs 1 and 2 absent.}
\\
\includegraphics[height=4.4cm]{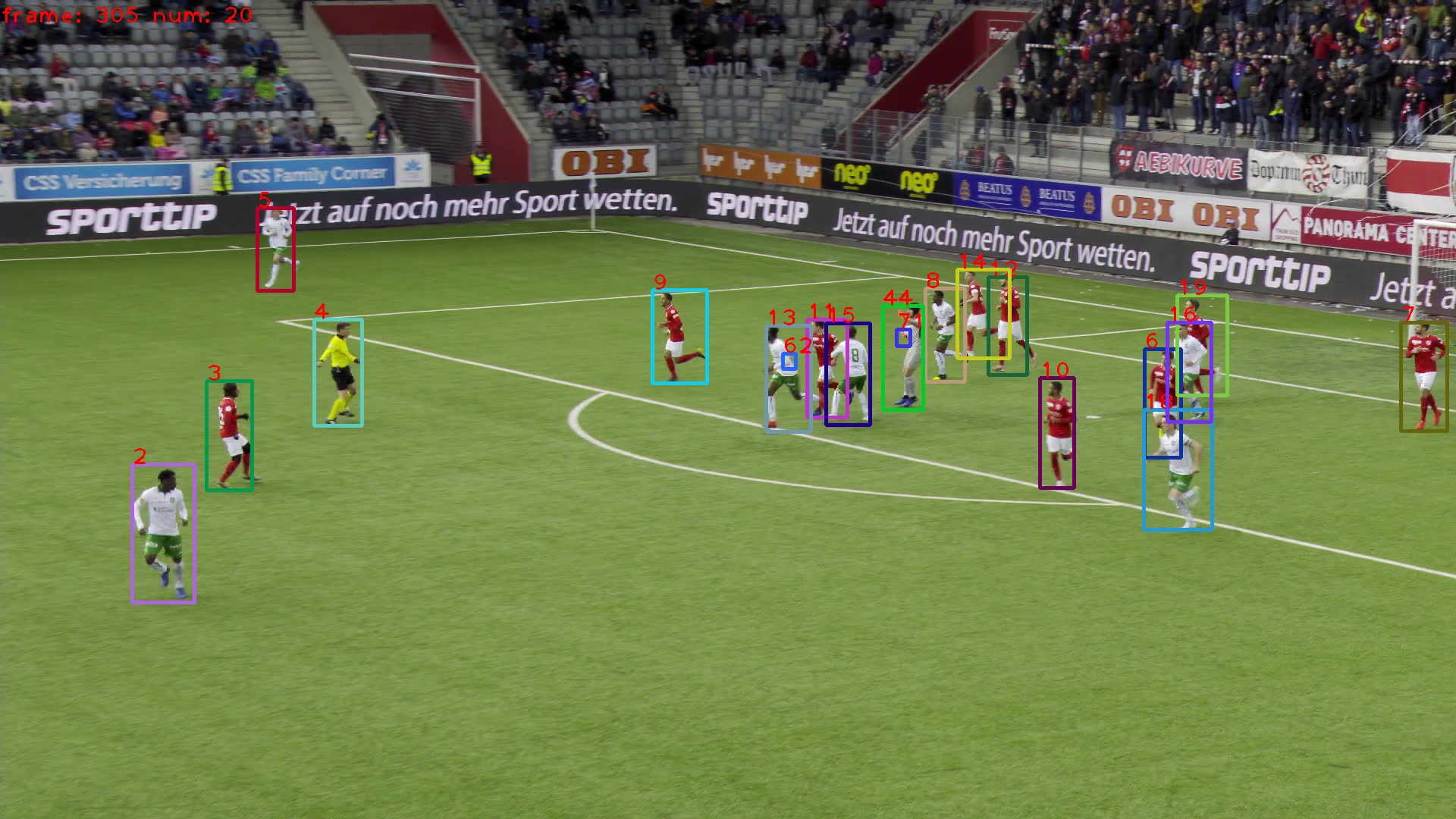}  &
\includegraphics[height=4.4cm]{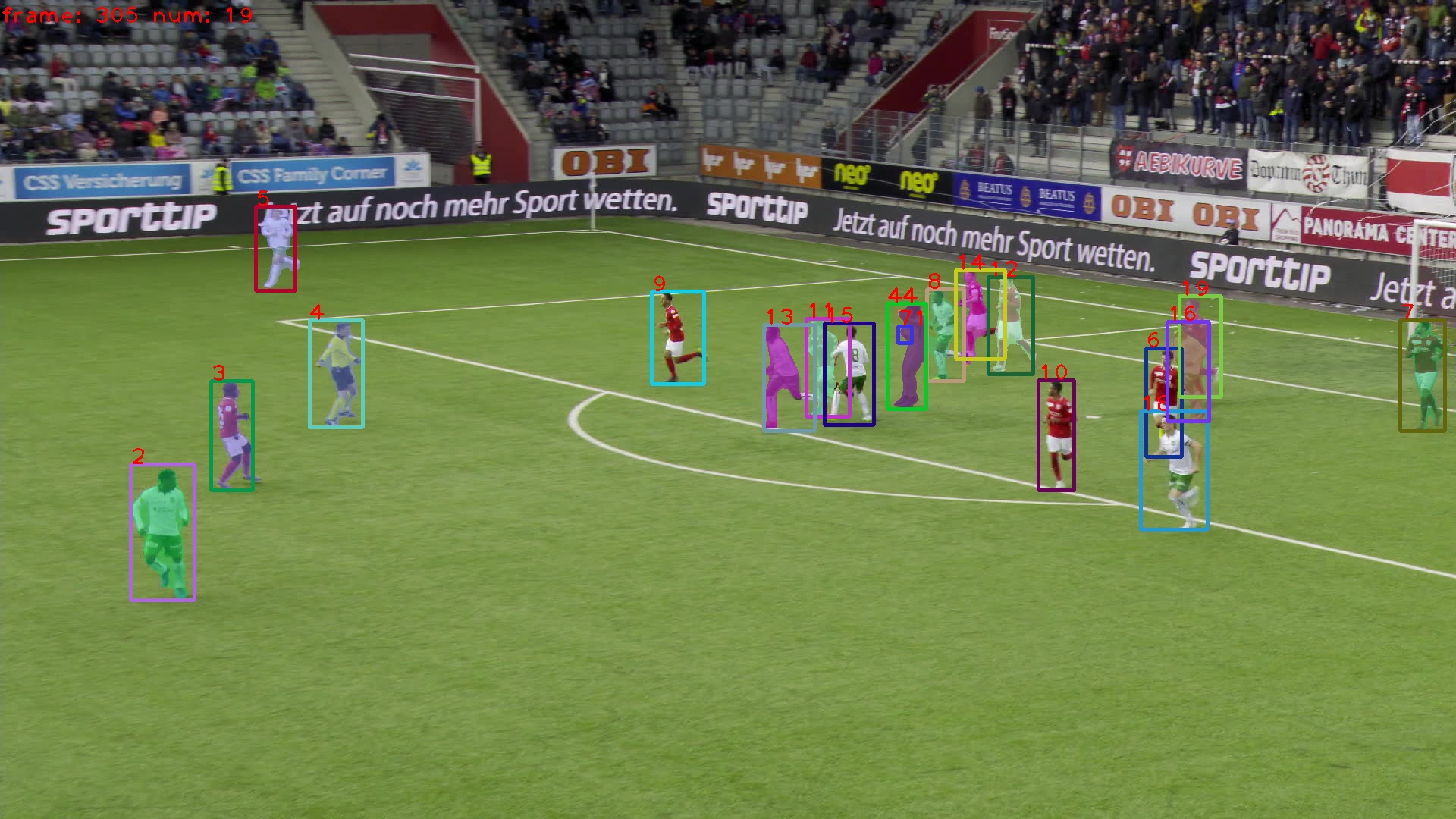}
\\
\multicolumn{2}{c}{Frame 305 - player with ID 2 reappears and is recognized.}
\\
\includegraphics[height=4.4cm]{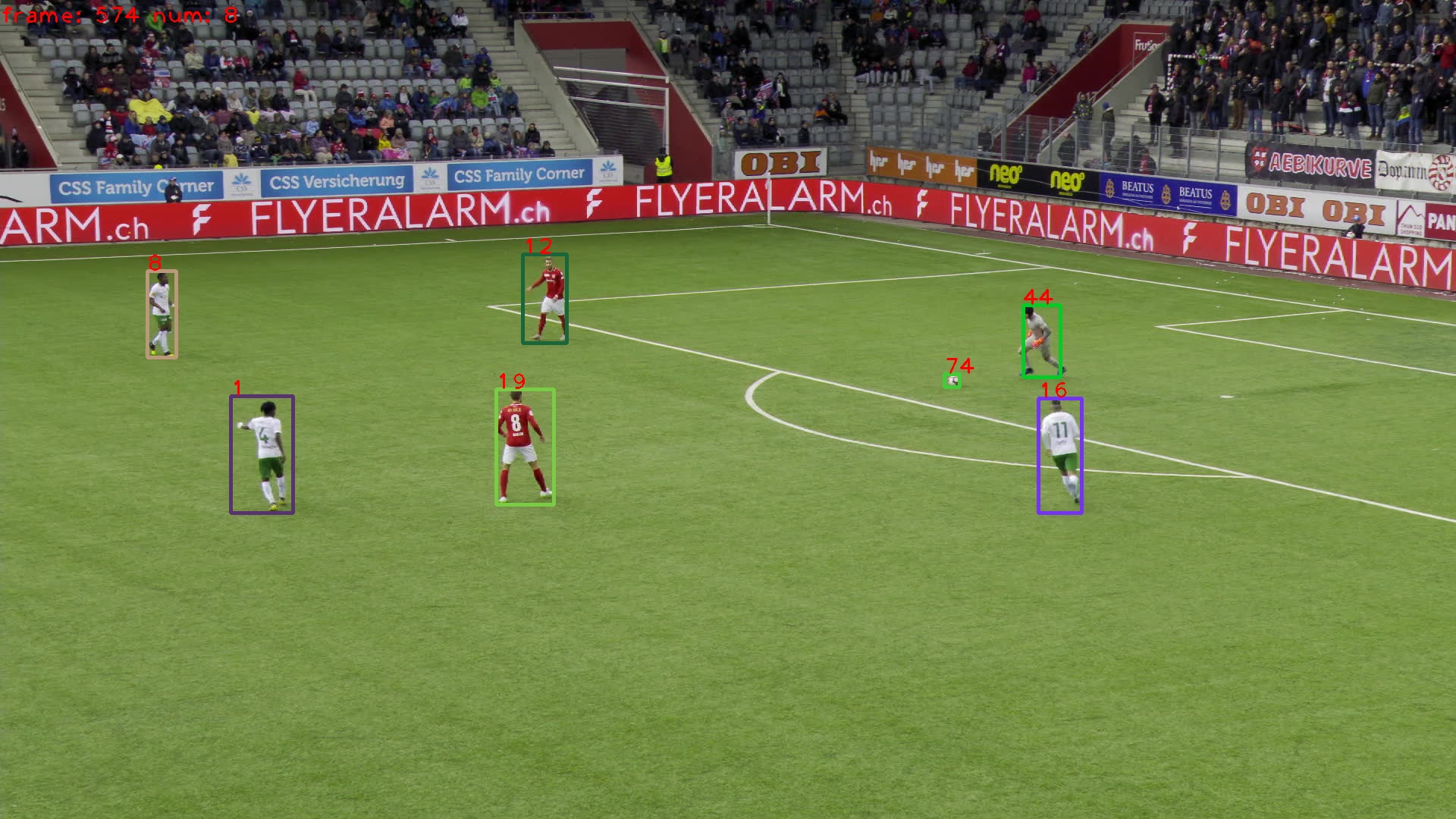}  &
\includegraphics[height=4.4cm]{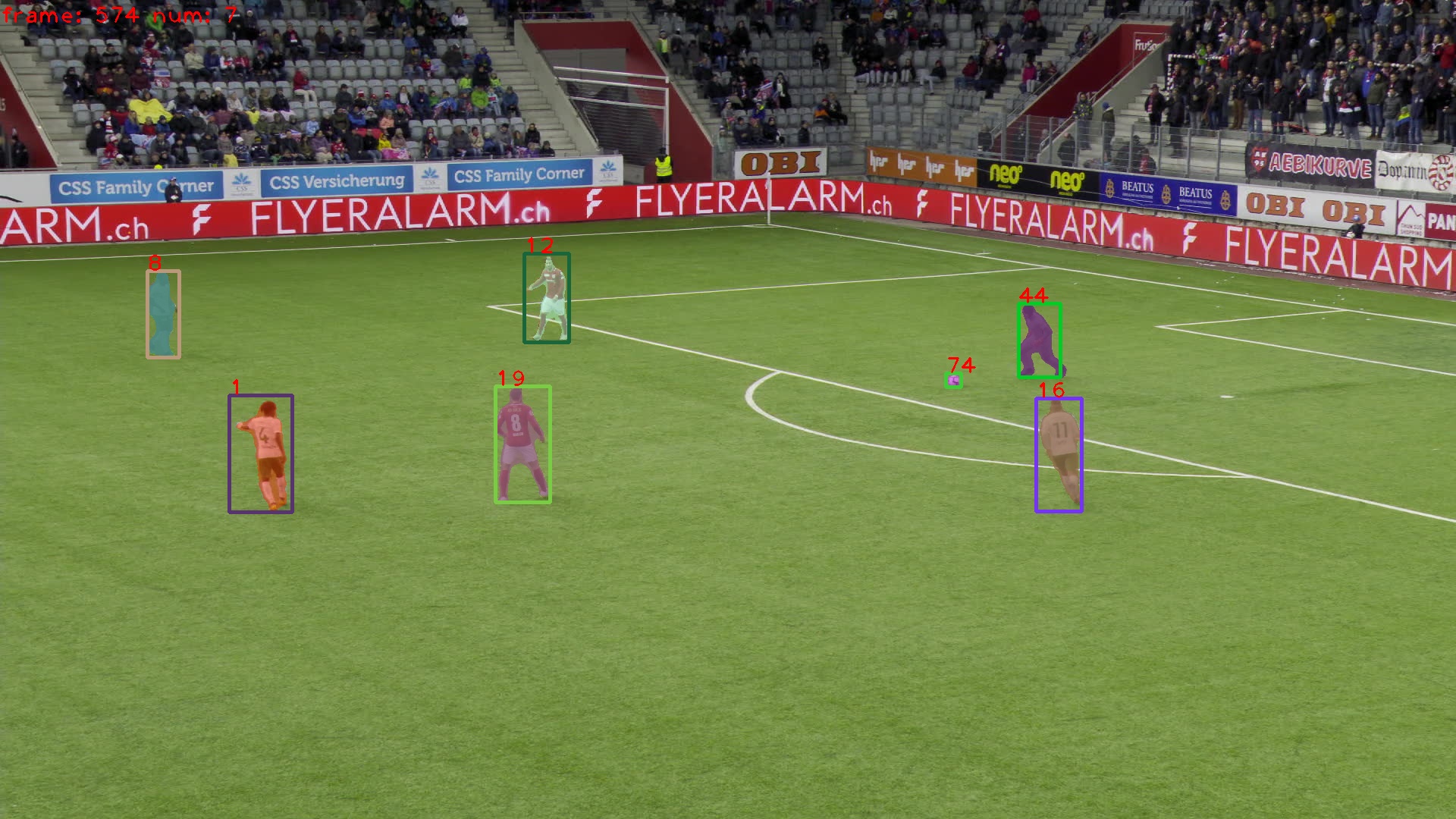}
\\
\multicolumn{2}{c}{Frame 574 - player with ID 1 reappears and is recognized.}
\end{tabular}
\caption{
Qualitative example of long-term identity preservation on SoccerNet-tracking. Players temporarily leave the field of view and reappear after several frames. McByte++ successfully reconnects the tracklets using online re-identification, preserving the original identity without introducing an ID switch. Bounding boxes are shown for clarity, while masks predicted by EdgeTAM are visualized only to illustrate spatial alignment and how they are used by McByte++. This example highlights the effectiveness of the online long-term identity association mechanism in realistic sports scenarios with frequent occlusions and camera motion.
}
\label{fig:full_football_comp}
\end{figure*}

Qualitative examples of long-term identity preservation are shown in Fig.~\ref{fig:full_football_comp}, where players temporarily leave the field of view and are correctly re-associated upon reappearance without identity switches. This illustrates the effectiveness of the online re-ID mechanism in maintaining long-term identities in sports scenarios.

\paragraph{Offline post-processing with GTA-link.}

For completeness, we also evaluate offline post-processing using GTA-link~\cite{gta_link_ref}, applied to the conditional CMC configuration ($ds=4$). On SoccerNet-tracking, GTA-link further improves HOTA to 88.6 and IDF1 to 87.2, achieving gains of +4.5 HOTA and +8.3 IDF1 over the green no-re-ID variant. On SportsMOT, GTA-link achieves HOTA 81.5 and IDF1 86.0. Runtime increases (FPS decreases) compared to pure online processing due to the additional offline step.

These results confirm that global post-processing can further enhance identity consistency. However, unlike our online re-ID mechanism, GTA-link operates offline and requires a complete tracker output as input.

\paragraph{SoccerNet-tracking Challenge 2023 split.}

We additionally report results on the SoccerNet-tracking Challenge 2023 split (Table~\ref{tab:soccernet_challenge_2023}), where detections are not provided. We use YOLOX~\cite{yolox_ref} trained on SportsMOT~\cite{sportsmot_ref}. Due to the unexpected unavailability of the official evaluation server, comparisons are limited to McByte variants.

Even under these constraints, McByte++ with online re-ID improves IDF1 from 74.1 to 78.6 and maintains competitive HOTA compared to the original McByte, while achieving more than a sevenfold speed increase (15.05 FPS vs. 1.46 FPS for the no-re-ID variant). These results further validate the robustness and practical efficiency of McByte++.

\begin{table}
  \centering
  {\small{
  \scalebox{0.9}{
  \begin{tabular}{lcccc}
    \toprule
    Method & HOTA$\uparrow$ & IDF1$\uparrow$ & MOTA$\uparrow$ & FPS$\uparrow$ \\
    \midrule
    \multicolumn{5}{c}{Online processing} \\
    % \midrule
    \rowcolor{lightyellow}
    McByte~\cite{MCBYTE_REF}                       & 64.1 & 76.5 & 81.8 & 1.46 \\
    \rowcolor{lightgreen}
    McByte++, no re-ID                                      & 62.4 & 74.1 & 81.7 & 15.05 \\
    \rowcolor{lightblue}
    McByte++, with re-ID                   & 64.3 & 78.6 & 81.8 & 11.13 \\  % buffered re-ID
    \midrule
    \multicolumn{5}{c}{Offline, with postprocessing} \\
    % \midrule
    \rowcolor{lightblue}
    McByte++ with GTA          & 65.7 & 80.7 & 81.7 & 10.06 \\
    \bottomrule
  \end{tabular}
  }
  }}
  \caption{
Results on SoccerNet-tracking Challenge 2023 split using YOLOX detections trained on SportsMOT. Due to unexpected unavailability of the official evaluation server during preparation of this work, comparisons are limited to McByte variants. McByte++ demonstrates improved identity preservation and substantially higher runtime efficiency compared to the original McByte.
}
  \label{tab:soccernet_challenge_2023}
\end{table}

\begin{table}
  \centering
  {\small{
  \scalebox{0.9}{
  \begin{tabular}{lcccc}
    \toprule
    Method & Train & HOTA$\uparrow$ & IDF1$\uparrow$ & MOTA$\uparrow$ \\
    \midrule
    \textcolor{black}{The same detections} & & & & \\
    ByteTrack~\cite{bt_ref}     & \ding{53} & 72.1 & 75.3 & 94.5 \\
    OC-SORT~\cite{ocsort_ref}   & \ding{53} & 82.0 & 76.3 & \textbf{98.3} \\
    \rowcolor{lightyellow}
    McByte~\cite{MCBYTE_REF}     & \ding{53} & 85.0 & 79.9 & 96.8 \\
    \rowcolor{lightblue}
    McByte++, online    & \ding{53} & \underline{87.5} & \underline{84.5} & \underline{97.1} \\
    \rowcolor{lightblue}
    McByte++, offline with GTA & \ding{53} & \textbf{88.6} & \textbf{87.2} & \underline{97.1} \\   
    \bottomrule
  \end{tabular}
  }
  }}
  \caption{
Comparison with state-of-the-art tracking methods on SoccerNet-tracking 2022 (test split). Oracle detections are used when provided. McByte++ operates without any detector retraining or dataset-specific tuning. HOTA and IDF1 are emphasized as primary identity and association metrics. The best results per metrics are in bold and second-best are underlined.
}
  \label{tab:sota_soccernet}
\end{table}

\begin{table}
  \centering
  {\small{
  \scalebox{0.9}{
  \begin{tabular}{lcccc}
    \toprule
    Method & Train & HOTA$\uparrow$ & IDF1$\uparrow$ & MOTA$\uparrow$ \\
    \midrule
    \textcolor{black}{Detections on their own} & & & & \\
    \textcolor{gray}{GTR~\cite{gtr_ref}} & \textcolor{gray}{\ding{51}} & \textcolor{gray}{54.5} & \textcolor{gray}{55.8} & \textcolor{gray}{67.9} \\
    \textcolor{gray}{CenterTrack~\cite{centertrack_ref}} & \textcolor{gray}{\ding{51}} & \textcolor{gray}{62.7} & \textcolor{gray}{60.0} & \textcolor{gray}{90.8} \\
    \textcolor{gray}{MeMOTR~\cite{memotr_ref}} & \textcolor{gray}{\ding{51}} & \textcolor{gray}{70.0} & \textcolor{gray}{71.4} & \textcolor{gray}{91.5} \\
    \textcolor{gray}{MOTIP~\cite{motip_ref}} & \textcolor{gray}{\ding{51}} & \textcolor{gray}{71.9} & \textcolor{gray}{75.0} & \textcolor{gray}{92.9} \\
    \textcolor{gray}{MotionTrack~\cite{motiontrack_ref}} & \textcolor{gray}{\ding{51}} & \textcolor{gray}{74.0} & \textcolor{gray}{74.0} & \textcolor{gray}{96.6} \\
    \textcolor{gray}{Deep-EIoU~\cite{deep_eiou_ref}} & \textcolor{gray}{\ding{51}} & \textcolor{gray}{77.2} & \textcolor{gray}{79.8} & \textcolor{gray}{96.3} \\
    \textcolor{gray}{SportMamba~\cite{sportmamba_ref}} & \textcolor{gray}{\ding{51}} & \textcolor{gray}{77.3} & \textcolor{gray}{77.7} & \textcolor{gray}{96.9} \\
    \midrule
    \textcolor{black}{The same detections} & & & & \\
   
    \textcolor{gray}{MixSORT~\cite{sportsmot_ref}} & \textcolor{gray}{\ding{51}} & \textcolor{gray}{74.1} & \textcolor{gray}{74.4} & \textcolor{gray}{96.5} \\
    \textcolor{gray}{DiffMOT~\cite{diffmot_ref}} & \textcolor{gray}{\ding{51}} & \textcolor{gray}{76.2} & \textcolor{gray}{76.1} & \textcolor{gray}{97.1} \\
    ByteTrack~\cite{bt_ref} & \ding{53} & 62.1 & 69.1 & 93.4 \\
    OC-SORT~\cite{ocsort_ref} & \ding{53} & 68.1 & 68.0 & 93.4 \\
    \rowcolor{lightyellow}
    McByte~\cite{MCBYTE_REF}       & \ding{53} & 76.9 & 77.5 & \textbf{97.2} \\
    \rowcolor{lightblue}
    McByte++, online      & \ding{53} & \underline{79.9} & \underline{83.6} & \underline{96.9} \\
    \rowcolor{lightblue}
    McByte++, offline with GTA & \ding{53} & \textbf{81.5} & \textbf{86.0} & 96.8 \\
    \bottomrule
  \end{tabular}
  }
  }}
  \caption{
Comparison with state-of-the-art tracking methods on SportsMOT (test split). Results are reported under identical detection settings when possible. McByte++ achieves the highest HOTA and IDF1 among multi-object tracking approaches while maintaining a fully training-free pipeline. The best results per metrics are in bold and second-best are underlined.
}
  \label{tab:sota_sportsmot}
\end{table}

\subsection{State-of-the-art comparison}

We compare McByte++ against representative recent multi-object tracking methods on SoccerNet-tracking 2022 and SportsMOT. Whenever possible, quantitative comparisons are performed under identical detection and evaluation settings to ensure a fair assessment of the association algorithm itself. For benchmarks where results under different detector configurations are available, we explicitly distinguish methods evaluated with their own detections from those evaluated using identical detections, since detector training, tracker supervision, adaptation procedures, and evaluation protocols can substantially influence the reported performance. Furthermore, recent publications do not always report evaluations against a common set of existing methods or benchmark baselines, making exhaustive cross-paper comparison inherently difficult. We therefore prioritize comparisons that can be interpreted under comparable evaluation settings. We additionally distinguish between training-based and training-free tracking approaches whenever appropriate. McByte++ operates without detector retraining, dataset-specific tuning, or model fine-tuning.

\paragraph{SoccerNet-tracking 2022.}

Table~\ref{tab:sota_soccernet} presents results on the SoccerNet-tracking 2022 test set. Since oracle detections are provided for this benchmark, MOTA values are primarily influenced by detection quality and are therefore less indicative of tracking performance differences. For this reason, we focus our analysis on HOTA and IDF1.

Among methods evaluated under identical detection settings, McByte++ with online re-identification achieves HOTA 87.5 and IDF1 84.5, outperforming ByteTrack (72.1 / 75.3) as well as the original McByte (85.0 / 79.9). The offline variant with GTA-link further increases performance to HOTA 88.6 and IDF1 87.2, achieving the highest identity preservation among compared methods.

Compared to the original McByte, McByte++ improves HOTnA by +2.5 and IDF1 by +4.6 in the online setting, and by +3.6 and +7.3 respectively when combined with offline post-processing. MOTA remains comparable across variants because all compared methods use the same detector outputs. Consequently, the proposed improvements primarily affect the association stage, leading to enhanced long-term identity consistency reflected by higher HOTA and IDF1 rather than changes in detection accuracy.

\paragraph{SportsMOT.}

Table~\ref{tab:sota_sportsmot} reports results on SportsMOT. This benchmark includes diverse sports scenes with strong camera motion and frequent occlusions. Under identical detection settings, McByte++ (online) achieves HOTA 79.9 and IDF1 83.6, surpassing ByteTrack (62.1 / 69.1) and the original McByte (76.9 / 77.5). The offline variant with GTA-link further improves performance to HOTA 81.5 and IDF1 86.0.

Compared to McByte, McByte++ improves HOTA by +3.0 and IDF1 by +6.1 in the online setting, and by +4.6 and +8.5 respectively with offline post-processing. MOTA differences remain small because the detector outputs are identical across the compared tracking variants. Since McByte++ improves tracklet association rather than object detection, its gains are naturally reflected in identity-aware metrics such as HOTA and IDF1.

Notably, McByte++ achieves the highest HOTA and IDF1 among the listed tracking-by-detection approaches under the same detection regime. Training-based methods evaluated with their own detections are included for reference; however, direct comparison is influenced by differences in detector training and supervision.

\paragraph{SoccerNet-tracking Challenge 2023.}

On the SoccerNet-tracking Challenge 2023 split (Table~\ref{tab:soccernet_challenge_2023}), detections are not provided and we use YOLOX trained on SportsMOT. Due to the unexpected unavailability of the official evaluation server during the preparation of this work, comparisons are limited to McByte variants. Even under these conditions, McByte++ consistently improves IDF1 and HOTA compared to the original McByte while maintaining substantially higher runtime efficiency.

\paragraph{Summary.}

Across benchmarks, McByte++ consistently improves identity preservation and long-term tracking performance compared to the original McByte and other training-free baselines, while maintaining a fully training-free pipeline. When combined with offline post-processing, McByte++ achieves the strongest identity metrics among compared approaches. These results demonstrate that robust long-term tracking in sports can be achieved without dataset-specific training or parameter tuning.

\begin{table}
  \centering
  {\small{
  \scalebox{0.9}{
  \begin{tabular}{lccc}
    \toprule
    Method & HOTA$\uparrow$ & IDF1$\uparrow$ & MOTA$\uparrow$ \\
    \midrule
    DEVA, original settings        & 39.3 & 37.3 & -109.6 \\
    DEVA, with YOLOX               & 42.4 & 42.1 & -57.0 \\
    \midrule
    Grounded SAM 2, original settings & 45.9 & 44.7 & -13.9 \\
    Grounded SAM 2, with YOLOX     & 66.1 & 70.2 & 91.4 \\
    \midrule
    MASA, original settings        & 39.4 & 35.7 & -27.2 \\
    MASA, with YOLOX               & 73.6 & 71.2 & 97.0 \\
    \midrule
    \rowcolor{lightyellow}
    McByte                  & 83.9 & 83.6 & \textbf{98.9} \\
    \rowcolor{lightblue}
    McByte++, online & \underline{85.5} & \underline{87.6} & \underline{98.7}\\
    \rowcolor{lightblue}
    McByte++, offline with GTA & \textbf{87.2} & \textbf{90.4} & 98.6\\
    \bottomrule
  \end{tabular}
  }
  }}

  \caption{
Comparison with segmentation-based tracking systems on SportsMOT \textit{validation set}. Results are reported using both original method settings and YOLOX detections trained on SportsMOT. McByte++ integrates lightweight mask propagation within a structured tracking-by-detection framework and achieves superior identity preservation compared to pure mask-based approaches. The original McByte is also listed for reference. The best results per metrics are in bold and second-best are underlined.
}
  \label{tab:deva_grsam2_masa_sportsmotval}
  \vspace*{-0.30cm}
\end{table}

\subsection{Comparison with mask-based methods}

To further evaluate the role of segmentation-based approaches in sports multi-object tracking, we compare McByte++ with representative mask-based tracking systems, including DEVA~\cite{deva_ref}, Grounded SAM 2~\cite{sam2_ref,gr_dino_ref}, and MASA~\cite{masa_ref}. Experiments are conducted on the SportsMOT \textit{validation set} using both the original settings of each method and YOLOX~\cite{yolox_ref} detections trained on SportsMOT for fair comparison. We also list the original McByte~\cite{MCBYTE_REF} for reference.

As shown in Table~\ref{tab:deva_grsam2_masa_sportsmotval}, McByte++ (both online and offline variants) outperforms all compared mask-based methods across HOTA and IDF1, while maintaining competitive MOTA. Compared to the original McByte, McByte++ further improves identity-related metrics, demonstrating the benefit of conditional camera motion compensation and long-term re-identification.

The particularly large improvement observed for MASA when replacing the original detections with YOLOX detections suggests a stronger dependence on detection quality than in the compared tracking-by-detection methods. Cleaner detections reduce false tracklet initialization and provide more reliable mask initialization, resulting in substantially improved tracking performance.

Mask-based tracking systems, while effective for single-object or short-term segmentation tasks, face challenges in the multi-object tracking setting. DEVA lacks an explicit tracklet management mechanism, which can lead to fragmented identities and unstable MOTA when identity switches accumulate. Grounded SAM 2 relies on segment-level tracking, but merging segments into coherent tracklets can be inconsistent in crowded sports scenes. MASA shows limitations under longer occlusions and rapid re-entries, occasionally failing to maintain identity continuity.

In contrast, McByte++ incorporates a lightweight segmentation module (EdgeTAM~\cite{edgetam_ref}) as an auxiliary cue within a structured tracking-by-detection framework. As demonstrated both in the original McByte~\cite{MCBYTE_REF} and in this work, segmentation masks alone are insufficient for robust multi-object tracking in sports. However, when integrated in a regulated manner together with bounding box association, camera motion compensation, and re-identification, mask information significantly enhances association robustness.

Overall, these results highlight that while pure mask-based tracking systems struggle in dense sports scenarios, the structured integration of lightweight mask propagation within McByte++ yields superior multi-object tracking performance, with McByte++ achieving the strongest identity preservation among all compared approaches.

\noindent\textbf{A note on SAM3.} It is worth noting that recent segmentation-based trackers such as SAM3~\cite{sam3_ref} represent powerful advances in video object tracking. However, we were unable to evaluate SAM3 within our experimental setup due to its substantial memory requirements, which exceeded the available GPU resources for processing long multi-object sports sequences. In practice, the high VRAM consumption of such models poses challenges for dense sports tracking scenarios with dozens of simultaneously visible subjects and extended video durations typical of MOT benchmarks. These practical constraints further motivate lightweight and structured tracking-by-detection frameworks, such as McByte++, which integrate segmentation cues efficiently while remaining scalable and deployable on standard hardware.

\section{Conclusion}

We presented McByte++, a training-free framework for efficient and long-term multi-object tracking in sports. By integrating lightweight mask propagation, conditional camera motion compensation, and online re-identification within a structured tracking-by-detection pipeline, McByte++ improves identity preservation while substantially increasing runtime efficiency compared to the original McByte.

Across SportsMOT and SoccerNet benchmarks, McByte++ consistently enhances HOTA and IDF1, demonstrating effective long-term identity association when players leave and re-enter the field of view. At the same time, replacing heavy segmentation components with a lightweight alternative and regulating camera motion compensation enables significant computational gains without compromising accuracy.

These results demonstrate that efficient long-term tracking does not require increasingly complex end-to-end architectures. Instead, McByte++ shows that selectively integrating complementary motion, spatial, and appearance cues within a structured tracking-by-detection framework can substantially improve long-term identity preservation while maintaining computational efficiency.

McByte++ provides a practical foundation for real-world sports analytics systems and illustrates the continued relevance of efficient, structured tracking frameworks in dynamic multi-object environments.

\section*{Acknowledgement}
This work has been supported by the French government, through the 3IA Cote d’Azur Investments in the project managed by the National Research Agency (ANR) with the reference number ANR-23-IACL-0001. The funder played no role in study design, data collection, analysis and interpretation of data, or the writing of this manuscript. 

This work was granted access to the HPC resources of IDRIS under the allocation 2025-AD011014370 made by GENCI. 

% \section*{Author Contributions}
% T.S. conceived the study, developed the methodology, implemented the code, performed the experiments, and wrote the main manuscript text. S.Y. contributed through discussions on the research direction and provided feedback on the manuscript. F.B. supervised the project, provided guidance on the research, and reviewed the manuscript. All authors reviewed and approved the final manuscript.

%%%%%%%%% REFERENCES
{
    \small

    \bibliographystyle{naturemag}
    \bibliography{main}
}

\end{document}